\documentclass[conference]{IEEEtran}
\IEEEoverridecommandlockouts

\usepackage{amsmath,graphicx,url,bbm,amssymb,algorithm,mathtools}
\usepackage{color}
\usepackage[noend]{algpseudocode}
\usepackage{subcaption}
\usepackage{blindtext}
\usepackage{algorithmicx}
\usepackage{algpseudocode}

\DeclarePairedDelimiter\floor{\lfloor}{\rfloor}
\DeclareMathOperator*{\argmin}{argmin} 
\newcommand\norm[1]{\left\lVert#1\right\rVert}

\makeatletter
\def\BState{\State\hskip-\ALG@thistlm}
\makeatother

\ifCLASSINFOpdf
\else
\fi
\begin{document}
%
\title{MV-YOLO: Motion Vector-aided Tracking by Semantic Object Detection\\
\thanks{This work was supported in part by the NSERC Grant RGPIN-2016-04590}}

\author{\IEEEauthorblockN{Saeed Ranjbar Alvar}
\IEEEauthorblockA{School of Engineering Science\\
Simon Fraser University\\
Burnaby, BC, Canada\\
Email: saeedr@sfu.ca}
\and
\IEEEauthorblockN{Ivan V. Baji\'{c}}
\IEEEauthorblockA{School of Engineering Science\\
Simon Fraser University\\
Burnaby, BC, Canada\\
Email: ibajic@ensc.sfu.ca}}

%


\maketitle

\begin{abstract}
Object tracking is the cornerstone of many visual analytics systems. While considerable progress has been made in this area in recent years, robust, efficient, and accurate tracking in real-world video remains a challenge. In this paper, we present a hybrid tracker that leverages motion information from the compressed video stream and a general-purpose semantic object detector acting on decoded frames to construct a fast and efficient tracking engine. 
The proposed approach is compared with several well-known recent trackers on the OTB  tracking dataset. The results indicate advantages of the proposed method in terms of speed and/or accuracy.
Other desirable features of the proposed method are its simplicity and deployment efficiency, which stems from the fact that it reuses 
the resources and information that may already exist in the system for other reasons. 
\end{abstract}
%
%

\begin{IEEEkeywords}
Object tracking, semantic tracking, motion vectors, region of interest
\end{IEEEkeywords}

\section{Introduction}
\label{sec:intro}
Visual object tracking is one of the fundamental tasks in computer vision, and the cornerstone of many visual analytics applications in video surveillance, smart homes/cities, independent living, human-computer interaction, and so on. Despite the significant advances in the performance of trackers in recent years, robust, efficient, and accurate tracking in real-world video remains a challenge. 

Existing tracking approaches can be classified in a number of ways. For the purposes of this study, a division in terms of the input data domain is useful: pixel domain, compressed domain, and hybrid. Pixel-domain trackers are the most abundant and the most well-studied in the literature. Many successful tracking approaches were developed in this group, such as those based on correlation filters (e.g.~\cite{DSST}) and those based on learned deep features (e.g.~\cite{CNN_SVM,gordon2017re3}). Advantages of this class of methods include their potential for high accuracy and the fact that they are video codec-agnostic. However, they tend to be resource intensive, because all pixel values need to be reconstructed, stored in memory, and processed.  


The second group of trackers operate on compressed-domain data, with only partial decoding of the video bit stream. Compressed-domain data carry valuable information that has been shown to be useful in many applications, such as face detection~\cite{MIPR18} and localization~\cite{ICASSP18}, motion segmentation~\cite{chen_tmm_2011}, and object segmentation and tracking~\cite{kb_tip_2013, moving_object_hevc}. The key insight from the studies in~\cite{chen_tmm_2011, kb_tip_2013, moving_object_hevc} is that motion vectors (MVs) and related coding syntax elements are good indicators of the movements of objects in the scene. Since this information already exists in the video bit stream, it seems natural to try to use it in tracking.
Advantages of compressed-domain trackers include efficiency and speed, since they avoid most of video decoding, pixel value storage and processing, and generally operate on less input data. Their downside is the dependence on the video coding method used to compress the video, as well as potentially lower accuracy, limited by the low resolution of the motion sampling grid: usually, a single MV is assigned to blocks/units of size $4 \times 4$ or larger. 

The third group of trackers are hybrid ones, trying to take advantage of both compressed and pixel-domain data. An example of such approach is~\cite{Hybrid_HEVC}, which performs tracking by combining MVs and block coding modes extracted from the High Efficiency Video Coding (HEVC) bit stream with the color information from the decoded Intra frames. 

The tracking method proposed in this work is also a hybrid one, combining decoded MVs with semantic object detection operating on fully decoded frames.  
The basic idea is that MVs, which already exist in the compressed video bitstream, are good enough to indicate the approximate location of the target object. Semantic object detector then refines the object's location by providing pixel-precision bounding box on the decoded frame. The idea of two-stage tracking (approximation followed by refinement) has also been advocated in two other recent works, Parallel Tracking and Verifying (PTAV)~\cite{PTAV} and ROLO~\cite{ROLO}. Both these approaches are pixel-domain trackers, while ours is the first hybrid one, to our knowledge. PTAV uses a fast but less accurate pixel-domain tracker in the first stage followed by a Siamese network based on VGGNet~\cite{VGG} for refinement in the second stage. In ROLO, the first stage approximation is given by the YOLO object detector~\cite{YOLO}, while the second stage refinement is provided by a Long Short-Term Memory (LSTM) network. 

The paper is organized as follows. In Section~\ref{sec:proposed}, we present the details of the proposed tracking method. In Section~\ref{sec:experimental} we describe the experiments, and discuss the results and comparisons with several representative trackers from the literature. Section \ref{sec:conclusion} concludes the paper.

\begin{figure*}[t]	
	\centering
	\centerline{\includegraphics[width=\textwidth]{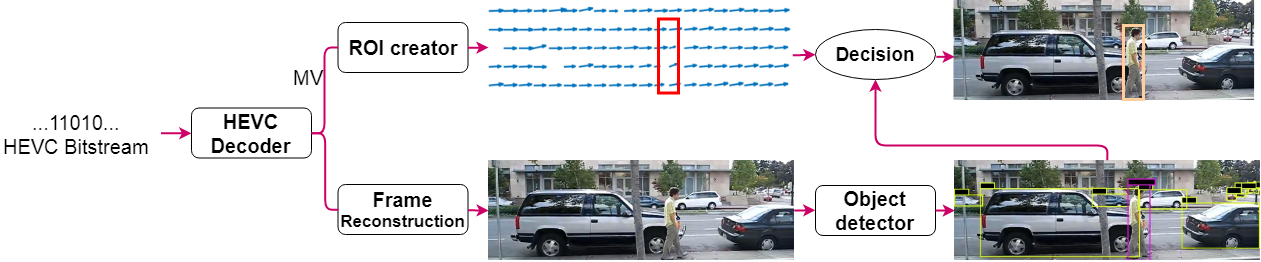}}
	\caption{An overview of the proposed tracking method}
	\label{fig:block_diagram}
\end{figure*} 

\section{Proposed Method}
\label{sec:proposed}
The proposed tracking framework is illustrated in Fig.~\ref{fig:block_diagram}. We refer to it as MV-aided YOLO, or MV-YOLO for short. Initially, approximate location of the target object is constructed based on MVs of the current inter-coded frame and the object's location in the previous frame. The constructed approximate location is referred to as the Region Of Interest (ROI). At the same time, decoded current frame is passed on to a semantic object detector (in our case YOLO), which detects locations of various objects in the frame. The ROI then helps decide which of these locations corresponds to the target object. Details are presented in the following subsections.

\subsection{ROI creation}
\label{subsec:ROI_creator}  
   
\begin{figure}[t]	
	\centering
	\centerline{\includegraphics[width=8.1 cm]{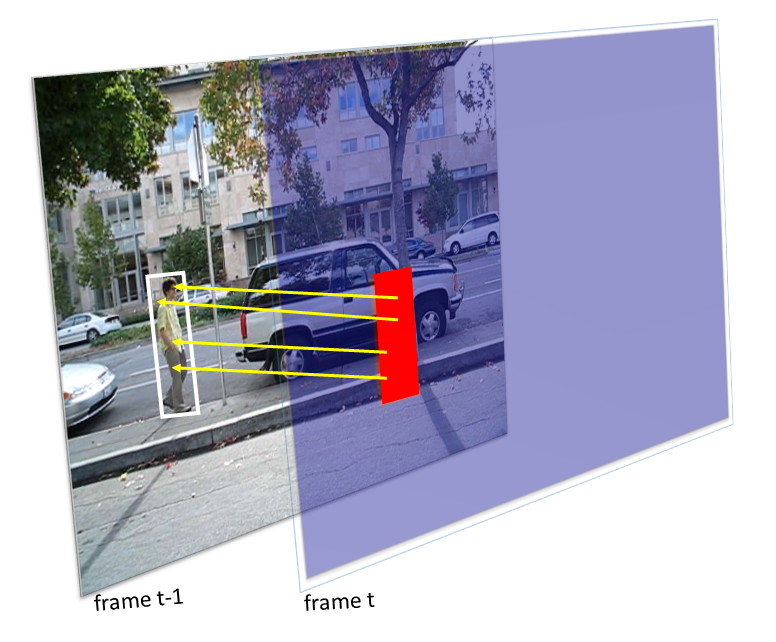}}
	\caption{An example of ROI creation}
	\label{fig:ROI}
\end{figure}

ROI creator uses the MVs from the HEVC bit stream to construct an approximate location of the target object in the current frame $t$, given the object's location in the previous frame $t-1$. The procedure is relatively simple.  
MVs of frame $t$ are read from the HEVC bit stream during frame decoding. The MV associated with a PU is assigned to all its pixels. Then, each pixel whose MV refers to the object's location in frame $t-1$ is labeled as ROI-pixel. Finally, ROI is selected as the smallest axis-aligned rectangle that includes all ROI-pixels. 
The process is illustrated in Fig.~\ref{fig:ROI}, where the ROI in frame $t$ is shown in red and several MVs from frame $t$ to frame $t-1$ are shown in yellow.
   
While the basic idea behind ROI creation is fairly intuitive, several technical challenges need to be resolved along the way. These include PUs without MVs (such as SKIP and intra-coded PUs), MVs pointing to frames other than $t-1$, and fractional precision MVs. Of these challenges, SKIP PUs are easiest to resolve. Since the SKIP mode indicates that the corresponding PU is almost exactly the same as the corresponding co-located region in the previous frame, zero MV is assigned to each SKIP PU.      
   
   
Assigning meaningful motion to intra-coded PUs is a bit more involved, since the fact that intra mode was chosen by the encoder is an indication that the underlying motion is too complicated to be taken advantage of by conventional motion compensation. For such PUs, we collect the MVs of all neighboring inter-coded PUs in the same Coding Tree Unit (CTU), and then apply Polar Vector Median (PVM)~\cite{kb_tip_2013} to come up with a suitable MV for such PUs.  
Specifically, let $V=(\mathbf{v}_1, \mathbf{v}_2, ... ,\mathbf{v}_n)$ be the list of MVs from neighboring PUs sorted according to their angle with respect to the horizontal axis. Then, a sub-list of $m = \floor*{(n+1)/2}$ consecutive vectors from $V$ is selected such that the sum of their angle differences is minimized. That is, the selected group is $(\mathbf{v}_k, \mathbf{v}_{k+1}, ..., \mathbf{v}_{k+m-1})$ where $k$ is chosen as
\begin{equation}
      k = \argmin_{j} \sum_{i=j}^{j+m-2} \left(\angle \mathbf{v}_{i+1} - \angle \mathbf{v}_{i} \right)
\end{equation}
Then the angle and the magnitude of the PVM vector $\widehat{\mathbf{v}}$ are set as:  
\begin{equation}
\begin{aligned}
     \angle \widehat{\mathbf{v}}  &= \textup{median}\left(\angle \mathbf{v}_k, \angle \mathbf{v}_{k+1}, ..., \angle \mathbf{v}_{k+m-1} \right) \\
       {\norm{\widehat{\mathbf{v}}}}_2 &= \textup{median}\left( {\norm{\widehat{\mathbf{v}}_1}_2}, {\norm{\widehat{\mathbf{v}}_2}_2}, ..., {\norm{\widehat{\mathbf{v}}_n}_2}\right)
   \label{eq:PVM}
\end{aligned}
\end{equation}

Finally, MVs in frame $t$ that point to frames other than $t-1$ are scaled (assuming continuity of motion) such that their scaled version points to frame $t-1$, similarly to~\cite{kb_tip_2013}. Components of fractional-precision MVs are rounded down to the nearest integer.

\subsection{Object detection}
\label{subsec:object_detector}
Semantic object detection refers to finding the locations of objects in the image and classifying them according to their type, e.g., human, car, dog, ... Any semantic object detector can be used in our proposed framework shown in Fig.~\ref{fig:block_diagram}. However, for the experiments, we chose three versions of the popular YOLO detector: YOLOv3~\cite{YOLOv3}, YOLOv2~\cite{YOLO9000}, and TinyYOLO, which is a simpler and faster (though less acurate) version of YOLOv2. 
   
Initial position of the object to be tracked is specified in the first frame of the sequence. Our tracker then tries to infer the object's class.\footnote{In some applications, object's class may be specified in the first frame, in which case it does not have to be inferred, and this step can be skipped.} This is done by running the object detector on the first five frames of the sequence. In each frame, the object detector outputs a number of boxes together with the object class with highest confidence for each box. In each frame, the detected box with the largest Intersection-Over-Union (IOU) with the specified location of the object is found, and the object class of that box is recorded. The most frequent among these object classes is the inferred class of the object to be tracked.  

After the object class is inferred, the frame at time $t$ is fed into the object detector and a set of $N$ boxes $\mathcal{B}=\{B_1,...,B_N\}$ are given as the output. These boxes carry the object location, object class, and confidence score. From these $N$ boxes, we eliminate all those whose class does not match the class of the object we are tracking. This way we end up with $M \leq N$ boxes, which we relabel as $\widehat{\mathcal{B}}=\{\widehat{B}_1,...,\widehat{B}_M\}$. These are used in the final decision stage, described below.
   
Note that the proposed tracking framework relies on semantics (i.e., object class) to eliminate some of the irrelevant objects/boxes in the frame. In principle, semantic information should help in difficult situations such as occlusion or multiple object tracking. However, even non-semantic detectors (those that do not output object class) could be used in our framework, but the accuracy would likely suffer due to a larger number of irrelevant boxes and the higher potential for making wrong decisions in the final stage. 

\subsection{Final box decision}
\label{sec:final_box}
After the object detector outputs a set of boxes $\widehat{B}$, the box corresponding to the target has to be identified. This is done with the help of the ROI found in the first stage. Among the boxes in $\widehat{\mathcal{B}}$, the one that has the highest IOU seems like a good candidate. However, even the highest IOU can be small. Hence, we also compare this highest IOU with an adaptive threshold in order to arrive at the final decision. Details are given in Algorithm~\ref{Alg:alg_1}.  

IOU between the ROI and the box $\widehat{B}_i \in \widehat{\mathcal{B}}$ is computed as 
\begin{equation}
\label{eq:IOU}
IOU(ROI,\widehat{B}_i) = \frac{\textup{Area} \left \{ROI \cap \widehat{B}_i \right \} }{\textup{Area} \left \{ROI \cup \widehat{B}_i 
\right \} }
\end{equation}
The adaptive threshold $T_{IOU}$ in Algorithm~\ref{Alg:alg_1} changes with respect to the IOU between the target and the ROI in the previous frame. Adaptation of this threshold (lines 10-18 in Algorithm~\ref{Alg:alg_1}) is designed to help with the cases where object detector fails to detect the target object, but instead detects the surrounding objects. It also helps in case of occlusion. In such cases, boxes produced by the object detector are not matching the target in the previous frame in terms of IOU (line 10 in Algorithm~\ref{Alg:alg_1}), so none of them are chosen, and instead the location of the target in the previous frame is taken as the final box $\widetilde{B}$ for the current frame (line 17 in Algorithm~\ref{Alg:alg_1}). But if the mismatch continues, the IOU acceptance threshold reduces ($T_{Reduction}$ increases in line 18 in Algorithm~\ref{Alg:alg_1}). Eventually, the lower IOU acceptance threshold (line 10 in Algorithm~\ref{Alg:alg_1}) will cause one of the detected boxes to be accepted as the the final box $\widetilde{B}$. 

\begin{algorithm}[tb]
\caption{Final box decision}
\begin{algorithmic}[1]
\Require ~~\ 
${Initial\_T}_{IOU}=0.7$  \Comment initial threshold 
\Require ~~\ 
$T_{IOU}=0.7$  \Comment adaptive threshold 
\Require ~~\ 
$T_{Reduction}=0.5$ \Comment threshold reduction
\Require ~~\ 
${\widehat{\mathcal{B}}}$  \Comment boxes found in the object detection stage
\Require ~~\ 
$M$  \Comment number of boxes in $\widehat{\mathcal{B}}$
\Require ~~\ 
$\mathcal{I}$ = \{\}\Comment IOU scores for the found boxes
\Ensure ~~\ 
${\widetilde{B}}$ \Comment final box

\If {$M == 0$}:
    \State No boxes in $\widehat{\mathcal{B}}$ $\implies$ Take the target location in frame \indent$t-1$ as  $\widetilde{B}$ 
\Else
    \For {$i=1$ to $M$}:
        \State Compute $IOU(ROI,\widehat{B}_i)$ from~(\ref{eq:IOU}) 
        \State Add $IOU(ROI,\widehat{B}_i)$ to $\mathcal{I}$
        \State $i \gets i+1$
    \EndFor
\State $j = \arg\max (\mathcal{I})$
\Comment{$\widehat{B}_j$ has largest IOU with ROI}


\State{Check validity of $\widehat{B}_j$}:
\If {$IOU(ROI,\widehat{B}_j) \geqslant (1-T_{Reduction})\cdot T_{IOU}$} 
    \State $\widetilde{B} \gets \widehat{B}_j$ 
    \Comment{Final box found} 
    \If { $IOU(ROI,\widehat{B}_j) > {Initial\_T}_{IOU}$ }
        \State $T_{IOU} \gets {Initial\_T}_{IOU}$
    \Else 
        \State $T_{IOU} \gets IOU(ROI,\widehat{B}_j)$
    \EndIf
\Else
    \State No suitable box is found $\implies$ Take the target \indent\indent location in frame $t-1$ as $\widetilde{B}$
    \State $T_{Reduction} \gets T_{Reduction}+0.2 $
\EndIf
\EndIf
\State \Return $\widetilde{B}$
\end{algorithmic}
\label{Alg:alg_1}
\end{algorithm}

\subsection{Summary}
\label{subsec:summary}
We now summarize several key features of the proposed tracking framework. 

\textbf{Compatibility with many object detectors}: 
One advantage of our tracking framework is that it is not crucially dependent on any particular object detector. While we use three versions of YOLO in our experiments for demonstration purposes, other detectors such as R-CNN~\cite{girshick14CVPR}, Fast R-CNN~\cite{girshickICCV15fastrcnn}, Faster R-CNN~\cite{renNIPS15fasterrcnn},  SSD~\cite{liu2016ssd}, and so on, can be used as well. 

\textbf{Resource sharing:} The object detector in our tracking framework may be used for other applications as well. For example, if the detector is placed in the cloud, other cloud services can use it for other purposes, such as object detection in user-supplied photos. This way, a single deep model can serve many applications.

\textbf{Data reuse:} In tracking, motion is usually one of the key challenges to conquer. But in our framework, motion is handled via MVs, which exist in the video bit stream anyway. This reuse of existing data speeds up the processing, and makes good engineering sense. 

\textbf{Robustness:} Other key challenges in tracking are appearance and scale changes. Many trackers try to model these explicitly. Our framework handles this problem by using an image-based object detector, which is not burdened by the memory of the object's appearance in the previous frames. As a result, the tracker is quite robust to appearance changes, as illustrated by an example in Fig.~\ref{fig:occlusion}(b). 
    
    
    

\section{Experimental Results}
\label{sec:experimental}
\subsection{Experimental settings}
\label{subsec:exp_settings}
A total of 30 sequences out of 100 sequences in OTB100 dataset~\cite{OTB100} were chosen for testing. These sequences contain object classes that are supported by YOLO. They are
listed in Table~\ref{tbl:sequences}. Test sequences were encoded using the HEVC reference software HM16.15~\cite{HM} with the configuration parameters in \texttt{encoder\_lowdelay\_P\_main.cfg}~\cite{hevc_ctc} and the Quantization Parameter (QP) set to 32. The motion vectors were then extracted from the compressed HEVC bit streams.

\begin{table}[t]
\centering
\caption{List of sequences used in the experiments}
\label{tbl:sequences}
\begin{tabular}{|c|c|c|c|c|}
\hline
 \multicolumn{5}{|c|}{Sequences} \\ \hline
Bird1 & BlurBody & BlurCar1 & BlurCar3 & Car4  \\ \hline
CarDark & CarScale & Couple & Dancer & Dancer2  \\ \hline
David3 & Diving & Dog & Girl2 & Gym    \\ \hline
Human2 & Human3 & Human6 & Human7 &  Human8  \\ \hline
Human9 & Jump & Singer1 & Singer2 & Skater    \\ \hline
Skater2 & Skating1 & Suv & Walking2 & Woman \\ \hline
\end{tabular}
\end{table}

The proposed tracking framework was compared against DSST~\cite{DSST} (the winner of the VOT 2014 challenge), CNN-SVM~\cite{CNN_SVM}, and Re3~\cite{gordon2017re3}. The latter two are representatives of the class of trackers based on deep neural networks that currently dominate this field. Within our framework, we used three versions of the YOLO object detector: YOLOv3~\cite{YOLOv3}, YOLOv2~\cite{YOLO9000}  and TinyYOLO~\cite{YOLO_site}, which is a simpler version of YOLOv2. The resulting trackers are referred to as MV-YOLOv3, MV-YOLOv2, and MV-TinyYOLO, respectively. The detection thresholds for YOLOv3 , YOLOv2 and TinyYOLO were set to 0.1, 0.1, and 0.03, respectively.

\subsection{Results}
\begin{figure}[t]
\begin{minipage}{\columnwidth}
\includegraphics[width=8.1 cm]{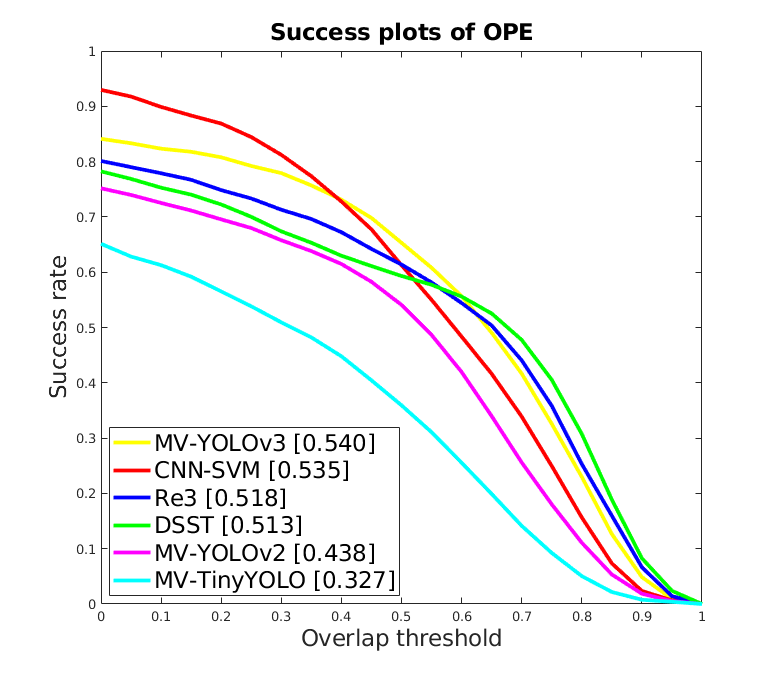}
\end{minipage}
\begin{minipage}{\columnwidth}
\includegraphics[width=8.1 cm]{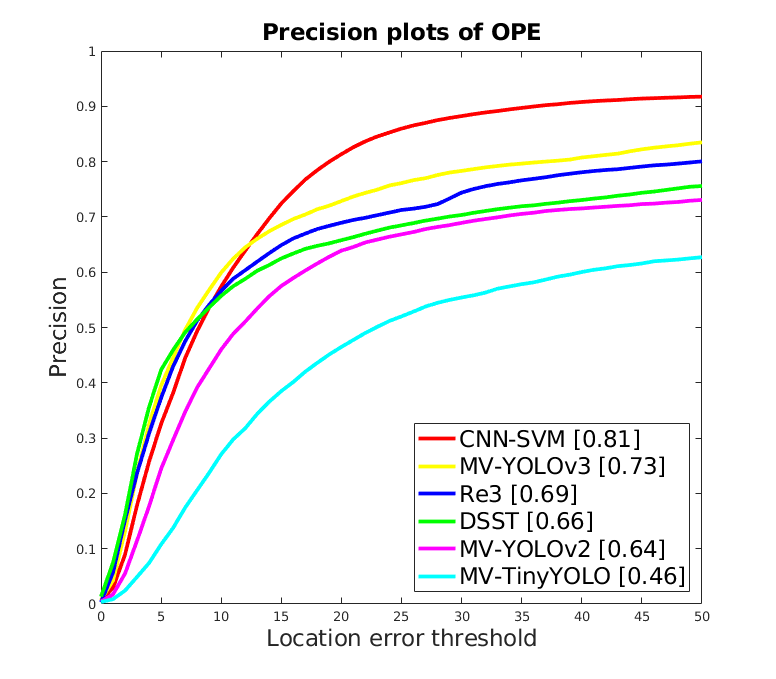}
\end{minipage}
\caption{Success and Precision curves. 
} 
\label{fig:results}
\end{figure}

To evaluate the trackers, one-pass evaluation (OPE)~\cite{OTB100} is performed. The Success and Precision plots~\cite{OTB100} are shown in Fig.~\ref{fig:results}. For each tracker, the Success curve is derived from the IOU of the predicted object box and the ground truth box, while the Precision curve represents the percentage of frames where average Euclidean distance between the centroids of the predicted box and the ground truth is less than a given threshold. 
In the Success plot (top graph in Fig.~\ref{fig:results}), the Area Under the Curve (AUC) is indicated in brackets next to each tracker in the legend. In the Precision plot (bottom graph), the numbers in the legend represent the percentage of the predicted boxes with centroids located within 20 pixels of the ground-truth centroid.



\begin{figure*}[t]	
	\centering
	\centerline{\includegraphics[width=\textwidth]{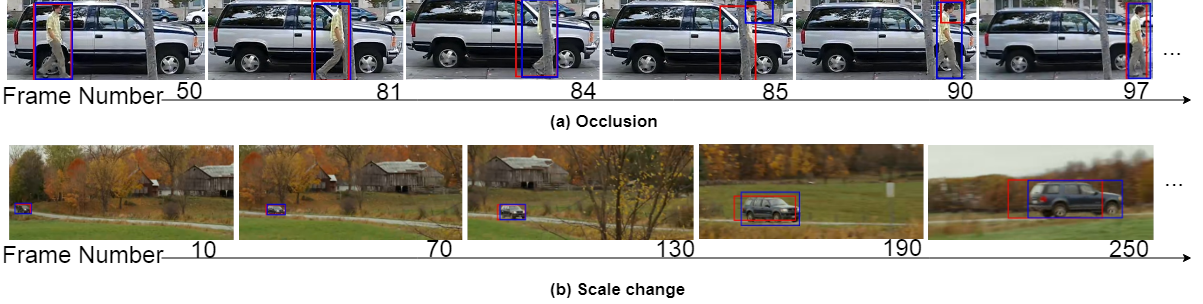}}
	\caption{The performance of the proposed method when (a)~occlusion or (b)~scale change occurs. The red box is the ROI derived from MVs (Section~\ref{subsec:ROI_creator}) and the blue box is the final predicted target location (Section~\ref{sec:final_box}).}
	\label{fig:occlusion}
\end{figure*} 
As seen in Fig.~\ref{fig:results}, the accuracy of the proposed tracking framework depends on the object detector employed. YOLOv3 leads to the best accuracy, followed by YOLOv2 and TinyYOLO. 
To further illustrate this point, Table~\ref{tbl:3_exp} shows Overlap Success Rate (OSR) and Distance Precision Rate (DPR)~\cite{OTB100} at thresholds of 0.5 and 20, respectively, for the three versions of the proposed tracker. Both DPR and OSR follow the trends in Fig.~\ref{fig:results}, indicating that MV-YOLOv3 is the most accurate of the three, and MV-TinyYOLO is the least accurate.   

However, speed results show the opposite trend. The last row of Table~\ref{tbl:3_exp} indicates the speed of the three trackers from the proposed framework. The tracker speed was computed as follows. We first measured the speed of ROI generation from MVs 
on a desktop machine with an Intel Corei7-6800K processor at 3.40 GHz, 128 GB RAM and a 12 GB Nvidia Titan X GPU. During this measuremen, any disk access time was ignored. Then we added this time to the object detection time reported 
on the official YOLO website~\cite{YOLO_site}. 
The inverse of the sum of these two times gives the speed in frames per second (fps) reported in the last row of Table~\ref{tbl:3_exp}. The fastest of the three trackers is MV-TinyYOLO at 88 fps, and the slowest is MV-YOLOv3 at 28 fps, which is still relatively fast. 

\begin{table}[t]
\centering
\caption{The performance and speed of the proposed tracking framework with three object detectors. 
}
\label{tbl:3_exp}
\begin{tabular}{|c|c|c|c|}
\hline
Metric & MV-YOLOv3 & MV-YOLOv2 & MV-TinyYOLO \\ \hline \hline
DPR (\%)      &  73       &    64    &    46        \\ \hline
OSR (\%)      &  65       &    54    &    36        \\ \hline
Speed (fps)   &  28       &    47    &    88         \\ \hline
\end{tabular}
\vspace{-0.2cm}
\end{table}

The Precision and Success results of DSST~\cite{DSST}, CNN-SVM~\cite{CNN_SVM} were taken from the official websites of each tracker. For Re3~\cite{gordon2017re3}, the authors of Re3 shared their results on OTB100 dataset with us. We see from Fig.~\ref{fig:results} that MV-YOLOv3 has the best average Success AUC among all tested trackers, while CNN-SVM has the best average Precision. In both precision and success results, MV-YOLOv3 is more accurate than Re3, which is encouraging. In turn Re3 is more accurate than DSST, which was the winning tracker in the VOT 2014 challenge. This illustrates the progress that has been made in the field in the last few years. All three versions of MV-YOLO are faster than CNN-SVM and DSST, but slower than Re3, according to the speed reported in the respective papers.

To further analyze the precision results of MV-YOLOv3 and CNN-SVM, the comparison of DPR (\%) at the threshold of 20 pixels is reported in Table~\ref{tbl:DPR_results} for each test sequence separately, where the better result is indicated in bold. Although CNN-SVM achieves higher DPR on average, MV-YOLOv3 outperforms CNN-SVM in almost half the test sequences. By examining the sequences in which MV-YOLOv3 has considerably lower performance such as Bird1 and CarDark, we found that problems arise in cases where an object of the same class as the one being tracked (e.g., bird or car) comes close to the ROI and the tracker accidentally ``latches on'' to it. Further work is needed to handle these situations, perhaps by incorporating object attributes into the tracking framework. Despite this, there are many sequences in which MV-YOLOv3 offers better DPR than the more complex and slower CNN-SVM.

\begin{table}[]
\centering
\caption{Comparison of DPR (\%) at 20 pixels distance in the tested sequences for MV-YOLOv3 and CNN-SVM~\cite{CNN_SVM}.}
\label{tbl:DPR_results}
\begin{tabular}{|c|c|c|}
\hline
Sequence & MV-YOLOV3         & CNN-SVM \\ \hline \hline
Bird1     & 14           & \textbf{37}      \\ \hline
BlurBody  & \textbf{86}  & 58      \\ \hline
BlurCar1  & 72           & \textbf{99}      \\ \hline
BlurCar3  & 68           & \textbf{99}      \\ \hline
Car4      & \textbf{100}          & \textbf{100}     \\ \hline
CarDark   & 9            & \textbf{100}     \\ \hline
CarScale  & \textbf{99}  & 70      \\ \hline
Couple    & 6            & \textbf{100}     \\ \hline
Dancer    & 43           & \textbf{96}      \\ \hline
Dancer2   & 91           & \textbf{97}      \\ \hline
David3    & 98           & \textbf{100}     \\ \hline
Diving    & \textbf{68}  & 43      \\ \hline
Dog       & 94           & \textbf{95}      \\ \hline
Girl2     & 34           & \textbf{94}      \\ \hline
Gym       & \textbf{98}  & 96      \\ \hline
Human2    & \textbf{74}  & 33      \\ \hline
Human3    & \textbf{93}  & 92      \\ \hline
Human6    & \textbf{97}  & 54      \\ \hline
Human7    & 90           & \textbf{99}      \\ \hline
Human8    & 98           & \textbf{100}     \\ \hline
Human9    & 89           & \textbf{100}     \\ \hline
Jump      & \textbf{66}  & 5       \\ \hline
Singer1   & 22           & \textbf{95}      \\ \hline
Singer2   & 5            & \textbf{83}      \\ \hline
Skater    & 78           & \textbf{90}      \\ \hline
Skater2   & \textbf{99}  & 81      \\ \hline
Skating1  & \textbf{99}  & 44      \\ \hline
Suv       & \textbf{95}  & 94      \\ \hline
Walking2  & \textbf{100} & 86      \\ \hline
Woman     & 98           & \textbf{100}     \\ \hline \hline
Average   & 73  & 81      \\ \hline
\end{tabular}
\vspace{-0.3cm}
\end{table}


Finally, some visual examples of the performance of MV-YOLOv3 are shown in Fig.~\ref{fig:occlusion}, where the red box indicates the ROI created from the MVs (Section~\ref{subsec:ROI_creator}) and the blue box shows the final predicted box (Section~\ref{sec:final_box}). Part (a) of the figure illustrates occlusion, where the pedestrian being tracked gets occluded by a tree trunk. Significant occlusion starts from frame 82 and continues until frame 85. In frames 82, 83 and 84, no box with significant overlap with ROI is found by the object detector, so the target box from frame 81 is chosen as the predicted target location (step 17 in  Algorithm~\ref{Alg:alg_1}) and the IOU acceptance threshold is reduced (step 18 in  Algorithm~\ref{Alg:alg_1}). In frame 85, only the head of the person is detected, and the IOU of the detected box and the ROI is relatively small. However, since the IOU threshold was reduced in frames 82, 83 and 84, the detected box is chosen as the target. The tracker locks on to the person and continues for a few frames with a small box tracking the head. Later, when the person is in full view and the object detector detects it fully, the tracker locks onto the person again (frame 97 and later). 

Part (b) of Fig.~\ref{fig:occlusion} shows the robustness of the proposed tracking framework to scale change. In frame 10, the car being tracked is small and located in the bottom-left part of the frame. Within the next 240 frames, the car moves towards right and towards the camera, while the camera itself also moves towards right. At frame 250, the car is about 15 times larger than it was in frame 10, and its appearance has changed: frame 10 was showing mostly the front view of the car, while frame 250 starts to show the rear view. Throughout these frames, the car is accurately tracked despite these appearance changes.  

\subsection{Final Remarks}
If the class of the object to be tracked is not supported by the object detector, there are two workarounds. The first is to fine-tune the detector (using transfer learning) for the desired object class. Alternatively, one could swithch to a generic object detector (e.g. ``objectness'').  

The proposed method relies on the MVs from the video bitstream, but is not necessarily dependent on MVs produced by an HEVC encoder. Since the earliest video coding standards, MVs were used to represent block-based transnational motion, and this is all that is required in the proposed framework. Our tracker only needs a hint of motion -- refinement of the object box is provided by the object detector.  

The performance of our tracking framework is highly dependant on the accuracy of the object detector, which in turn depends on the input image quality. Unfortunately, video sequences in the OTB100 dataset are stored frame-by-frame as JPEG images, and the quality of these JPEG images is not particularly good. In certain cases, coding artifacts can easily be seen. To obtain motion vectors for our tests, we had to further encode and decode these using HEVC, which has created additional artifacts. 
This has caused the object detector to miss the target object or wrongly classify objects in some cases, which negatively impacted our results. Other trackers in the study were not fed with HEVC-coded frames, so their performance was not affected by additional HEVC coding. 

\section{Conclusion}
In this paper we proposed MV-YOLO, a novel tracking framework that incorporates data reuse from the compressed video bit stream and semantic object detection. Based on the MVs extracted during the video decoding process, a ROI for the target object is created in the current frame. Then the output of a semantic object detector is used to more precisely localize the target object with the help of the ROI. 

The experiments show that MV-YOLO is a fast and robust tracking framework. The accuracy and speed of MV-YOLO depends on the particular object detector being used. However, even the slowest version we tested was reasonably fast at 28 fps, while its accuracy was comparable to the recent trackers based on deep models. 

In the present study, we examined only single object tracking. However, the MV-YOLO framework contains all the ingredients to support multiple object tracking as well. This is a topic for future research. 
\label{sec:conclusion}

\bibliographystyle{IEEEbib}
\bibliography{refs}
\end{document}